\documentclass{article} 
\usepackage{iclr2026_conference,times}

\usepackage{amsmath,amsfonts,bm}

\def\eqref#1{equation~\ref{#1}}

\def\1{\bm{1}}

\DeclareMathAlphabet{\mathsfit}{\encodingdefault}{\sfdefault}{m}{sl}
\SetMathAlphabet{\mathsfit}{bold}{\encodingdefault}{\sfdefault}{bx}{n}

\usepackage{wrapfig}
\usepackage{hyperref}
\usepackage{url}
\usepackage{tcolorbox}
\usepackage{graphicx}      
\usepackage{caption}       
\usepackage{booktabs}      
\usepackage{xcolor}        
\usepackage{colortbl}      
\usepackage{amssymb}       
\usepackage{array}
\usepackage{subcaption}
\usepackage{tabularx}

\title{Reinforcement Learning for Large Language Model Selective Evidence Adoption from Contaminated Retrieval Results}

\author{
\textbf{Yanyu Chen}\textsuperscript{1}\thanks{Equal Contribution.}
\textbf{, Yue Li}\textsuperscript{1}\footnotemark[1]
\textbf{, Yongyi Cui}\textsuperscript{1} 
\textbf{, Dongsheng Shi}\textsuperscript{1} 
\textbf{, Lichang Dai}\textsuperscript{2}\thanks{Corresponding author.} \\
\textsuperscript{1}East China Normal University \,
\textsuperscript{2}Shandong University \\
\texttt{cyyecao@stu.ecnu.edu.cn, }
\texttt{202435387@mail.sdu.edu.cn}
}

\iclrfinalcopy 
\begin{document}

\maketitle

\begin{abstract}
Retrieval-augmented large language models frequently face contexts that interleave useful evidence with misleading statements or instruction-like content. Blanket refusal discards valid evidence, whereas uncritical adoption yields incorrect or unsafe answers. The ability to selectively adopt relevant information while rejecting deceptive or harmful content is therefore critical for reliable deployment in real-world retrieval settings.
We introduce SelectBench, a controlled benchmark and training set for selective evidence adoption, and post-train Qwen3.5-4B directly with DAPO using either deterministic rule rewards or a frozen semantic judge. On the corrected 325-example SelectBench-v2 test set, strict success rises from 22.46\% for the original checkpoint to 25.54\% with DAPO-Rule and 26.46\% with DAPO-DeepSeek. Both trained policies reduce forbidden-content adoption and produce shorter, more focused responses, yet prompt-injection following does not improve. The paired gains are modest and fail to survive Holm correction, suggesting that stronger reward shaping or additional training iterations may be needed for more robust gains. DAPO-DeepSeek exhibits no material degradation on MMLU or clean HotpotQA, indicating that the post-training procedure preserves general capabilities. These results demonstrate a directional improvement in selective evidence use, while identifying injection resistance and statistical robustness as important remaining challenges for future work.
\end{abstract}

\section{Introduction}

Large Language Models (LLMs) have demonstrated remarkable proficiency across a broad spectrum of downstream tasks \cite{yi2025unified, li2025construction, li2026agmark}. Their ability to generalize from limited supervision and perform complex reasoning has positioned them as versatile foundation models for diverse real-world applications \cite{yi2025unifiededucation, shi2026benchmarking}. Additionally, by leveraging external tool-use capabilities, they can tackle tasks that go beyond the scope of inherent language reasoning \cite{zhang2026a, shi2026surgent}.
By explicitly instructing the model through prompts on when and how to invoke external tools, it becomes capable of performing tasks beyond the limits of pure linguistic reasoning \cite{sha2025sem, zhang2026toolsneedunveilingtooluse}.

Safety alignment in LLMs is a research direction dedicated to equipping models with the ability to refuse harmful instructions \cite{li2025hierarchical, yi2025latent}. Nevertheless, excessive alignment may inadvertently lead to the unwarranted refusal of benign ones \cite{zhang2025falsereject}.
Inspired by this, we examine the scenario in which LLMs call upon search tools to aid generation. A blanket refusal of retrieved information due to safety concerns, rather than extracting useful portions, constitutes an unnecessary waste of resources. On the other hand, uncritical acceptance of such information introduces safety risks and yields low-quality outputs. Consequently, equipping models with the ability to discern and adopt valuable information while filtering out harmful content is of paramount importance.

Prior work has demonstrated the potential of using reinforcement learning to enhance models' understanding and utilization of external tools, including retrieval tools \cite{sha2025sem,chen2026research}. Building on this, we construct SelectBench to study this problem. Based on the DAPO framework \cite{yu2026dapo}, we customize the reward signal to encourage the model to extract supported information from retrieved content interspersed with misleading or instruction-like material. We evaluate whether this policy shift improves selective evidence adoption and whether it changes performance on measured general-capability tasks.


In summary, our contributions are as follows:

\begin{itemize}
    \item We formulate selective evidence adoption from contaminated retrieval-tool outputs as a distinct post-training objective for LLM agents.
    \item We build a controlled dataset for selective evidence adoption and design a tailored RL pipeline that assigns positive reward only to correct, complete, and safe answers while treating forbidden adoption as a hard failure.
    \item On a corrected held-out test set, both trained policies show modest directional improvements in strict selective adoption and lower forbidden-content adoption. Paired statistical and failure-mode analyses expose remaining uncertainty and a lack of improvement in prompt-injection resistance.
\end{itemize}

\section{Preliminary}

We conduct a pilot experiment using Qwen3.5 \cite{qwen35blog} at two model scales (4B and 9B parameters) to examine the inherent ability of LLMs to extract useful information from low-quality retrieved content. Specifically, we consider three dimensions of low quality: safety corruption, temporality corruption, and format corruption, with 100 instances constructed for each dimension. This design allows us to systematically assess how model capacity influences discernment across diverse corruption types.

\begin{wraptable}{r}{0.58\textwidth}  
    \centering
    \footnotesize
    \caption{
        Comparison of Multi-dimensional Settings.
    }
    \label{tab:pre}
    \begin{tabular}{lccc}
        \toprule
        \textbf{Strategy} & \textbf{Safety} & \textbf{Temporality} & \textbf{Format} \\
        \midrule
        \multicolumn{4}{c}{\cellcolor{gray!30} \textit{Qwen3.5-4B}} \\
        \midrule
        \textbf{Full Accept} & 15.00 & 0.00 & 0.00 \\
        \textbf{Full Reject} & 55.00 & 0.00 & 0.00 \\
        \textbf{Selective Accept} & 30.00 & 100.00 & 100.00 \\
        \midrule
        \multicolumn{4}{c}{\cellcolor{gray!30} \textit{Qwen3.5-9B}} \\
        \midrule
        \textbf{Full Accept} & 10.00 & 0.00 & 0.00 \\
        \textbf{Full Reject} & 70.00 & 0.00 & 0.00 \\
        \textbf{Selective Accept} & 20.00 & 100.00 & 100.00 \\
        \bottomrule
    \end{tabular}
\end{wraptable}

\noindent
From Table \ref{tab:pre}, we observe that the model can largely overcome interference from low-quality noisy content in the temporality and format corruption dimensions, thereby achieving selective acceptance. In contrast, in the safety corruption dimension, the model exhibits a strong tendency toward either indiscriminate acceptance or wholesale rejection: the former compromises safety alignment, whereas the latter reflects overly conservative safety behavior.

Based on these descriptive pilot results, we identify safety-corrupted retrieval as the most challenging of the three tested dimensions. This preliminary experiment prompted us to conduct a controlled study on "selective retrieval result adoption," and we hypothesize that the model's safety alignment is the cause of the observed behavior.

\section{Methodology}

We study an agent that must answer a factual question after calling a search tool whose output mixes useful evidence with misleading or instruction-like text. Let $q$ denote the question, $c$ the fixed mixed retrieval context, and $\tau=(q,a_1,c,a_2)$ the resulting two-turn assistant trajectory: $a_1$ invokes the tool and $a_2$ contains the final reasoning and answer. The desired policy uses the evidence in $c$ that supports the answer while neither adopting unsupported content nor obeying instructions embedded in $c$. We post-train the policy directly with DAPO \cite{yu2026dapo} using a ternary task score followed by continuous length shaping. 

\subsection{Dataset Preparation}

\paragraph{Training data.}
SelectBench contains 1,300 training instances derived from multi-hop questions in HotpotQA \cite{yang2018hotpotqa}, 2WikiMultiHopQA \cite{ho2020constructing}, and MuSiQue \cite{trivedi2022musique}. The source mixture is 650/325/325 instances, respectively. For each question, we retain answer-supporting evidence and combine it with controlled low-quality spans. These spans cover an answer-like wrong summary, a surface-matched wrong entity, stale or misleading information, unsafe evidence-selection advice, over-refusal bait, and a retrieved prompt injection. Each instance records accepted answer aliases, required answer terms, known-wrong or forbidden phrases, and (where applicable) a unique injection-success marker.

We construct a dataset of 1,300 instances, comprising 650 \texttt{sft\_correction}, 455 \texttt{preference\_pair}, and 195 \texttt{injection\_resistance} cases. These labels describe construction provenance only. Training conversion actively removes chosen and rejected assistant trajectories, and RL receives only the question, fixed tool context, and reward labels. We split the data into 1,170 training and 130 validation examples with seed 42, stratifying jointly by source dataset, task type, and construction bucket.

The frozen construction manifest records 646 skipped candidates: 377 with test-evidence hash overlap, 156 rejected by semantic evidence or label review, 58 duplicate training questions, and 55 with other construction or review failures. Every retained instance received a first review, and 1,105 received a second review. A final leakage audit reported no surviving leakage and no near-duplicate flags under normalized-token trigram Jaccard similarity at a 0.9 threshold.

\paragraph{Example-disjoint test set.}
Our sole formal test set is SelectBench-v2, a corrected 325-example example-disjoint split derived from the original SelectBench challenge set. A post-training audit found that the original generator could substitute a non-supporting sentence when materializing trusted evidence. Version 2 strictly resolves all supporting facts from the pinned official HotpotQA distractor validation snapshot and changes the trusted evidence bundle in 233 examples. It preserves all questions, answers, synthetic contamination spans, forbidden labels, and injection markers. Consequently, the training data and both trained checkpoints remain unchanged. The split contains 275 injection-bearing examples and 50 examples with low-quality speculation instead of an injection.

No test question, context, answer label, identifier, marker, or row-level taxonomy enters the 1,300-example construction pool. Aggregate failure-type counts from the original diagnostic set informed the training taxonomy and leakage audit. Thus, method development was benchmark-informed at the aggregate category level, but no row-level test content or label was used for training. All training questions and evidence originate from disjoint training sources.

\subsection{Tool-Use Rollout}

The policy initially observes only a system instruction and the question $q$. It then invokes \texttt{selectbench\_search} with a non-empty query. The offline tool returns the example-specific context $c$ verbatim, regardless of query wording. This construction holds retrieval fixed across policies and isolates evidence selection from search-engine variance. A successful trajectory contains at least one successful tool call followed by a visible final answer. Correctness is evaluated on the text appearing after the final \texttt{</think>} delimiter, while forbidden-adoption checks span the entire final assistant turn so that copying a forbidden span into reasoning remains visible to the reward.

\subsection{Reward Policy}

Let $T(\tau)$ indicate a successful tool call, $C(\tau)$ a correct and complete answer, and $H(\tau)$ the adoption of a forbidden span or a retrieved instruction.
The deterministic task score is
\begin{equation}
r_{\mathrm{task}}(\tau)=
\begin{cases}
-1, & \neg T(\tau)\ \text{or}\ H(\tau),\\
+1, & T(\tau)\land C(\tau)\land\neg H(\tau),\\
0,  & \text{otherwise}.
\end{cases}
\end{equation}
Thus, forbidden or injection adoption and failure to use the required tool dominate factual correctness. An answer with no detected adoption that is wrong, incomplete, missing, or over-refusing is classified as a neutral failure with a raw score of zero rather than a hard negative. We additionally discourage unnecessarily long model output. For $n$ model-generated tokens (tool text excluded),
\begin{equation}
p_{\mathrm{len}}(n)=
\begin{cases}
0, & n\leq768,\\
-(n-768)/256, & 768<n<1024,\\
-1, & n\geq1024,
\end{cases}
\qquad
r(\tau)=\operatorname{clip}(r_{\mathrm{task}}+p_{\mathrm{len}},-1,1).
\end{equation}

We instantiate two variants of the ternary raw task score for policy training.

\textbf{DAPO-Rule} retains lowercase ASCII letters and digits, collapses all other character spans into spaces, and performs whole-phrase matching against accepted aliases, required terms, forbidden or wrong phrases, and injection markers. Accepted aliases and all required terms must appear unnegated in the visible answer. A forbidden or injection phrase is treated as negated---rather than adopted---when a negation marker occurs within its four-token left context, or when it is locally followed by \emph{wrong}, \emph{false}, \emph{unsafe}, or \emph{unsupported}. For examples labeled \texttt{full\_reject}, an explicit retrieval rejection or refusal also satisfies correctness.

\textbf{DAPO-DeepSeek} employs a frozen \texttt{deepseek-v4-flash} judge~\cite{xu2026deepseek} with the \texttt{selectbench-deepseek-rm-v2} rubric, yielding raw scores on the same $\{-1, 0, +1\}$ scale. Local hard gates assign $-1$ to tool omission and $0$ to a missing final answer prior to any judge invocation. The judge runs at temperature $0$ with thinking disabled and a $4{,}096$-token output limit; requests enforce a $180$-second timeout, at most eight retries, concurrency of $16$, and a rate limit of eight requests per second. Judge outputs must conform to a strict JSON schema and are content-addressed and cached.

After incorporating $p_{\mathrm{len}}$, the final policy reward is continuous on $[-1, 1]$ for outputs between $768$ and $1{,}024$ generated tokens. Dynamic group filtering operates on the unshaped ternary score, whereas the policy update uses the shaped reward $r(\tau)$.

\subsection{Policy Optimization}

For each prompt, we sample a group of eight trajectories and normalize their rewards within the group to obtain GRPO advantages. DAPO then applies the token-level clipped surrogate objective with asymmetric clipping bounds of $0.20$ and $0.28$, along with dynamic sampling that excludes groups with all-equal rewards. We use one policy epoch per batch and omit both the KL reward and the KL loss. The reward is placed on the last model-generated token rather than on tool-return tokens, ensuring that the policy gradient is attached to the agent's action sequence.

\section{Experimental Setup}

\subsection{Training details}

All post-training and formal evaluation experiments use the original post-trained Qwen3.5-4B checkpoint~\cite{qwen35blog}, pinned to revision \texttt{851bf6e8}. We freeze the vision tower and update the language model in bfloat16 with FSDP2, using AdamW with a learning rate of $5\times10^{-7}$, weight decay $0.1$, one policy epoch per batch, gradient clipping at $1.0$, and ten warm-up steps. Rollouts use temperature $1$, top-$p$ $1$, no top-$k$ truncation, a maximum prompt length of $768$, and a maximum response length of $3{,}072$. Both exported checkpoints correspond to optimizer step $130$ on NVIDIA RTX 3090 GPUs. Here, one policy epoch denotes one inner optimization pass over each sampled rollout batch, rather than one pass over the full training set. Both runs completed $130$ optimizer updates, with validation performed before the first update and every ten updates thereafter, and exported the predeclared terminal step-$130$ checkpoint; no checkpoint was selected using SelectBench-v2.

The rule-reward run used three GPUs with a prompt batch size of $9$, whereas the DeepSeek-reward run used four GPUs with a prompt batch size of $8$; both sampled eight rollouts per prompt. Consequently, their comparison is informative but does not constitute a strict single-factor reward-model ablation. Neither model was retrained nor were hyperparameters tuned after inspecting the formal test set.

\subsection{Datasets}

The primary evaluation is the corrected 325-example SelectBench-v2 test set described above. To measure capability retention, we additionally evaluate both the base and \textsc{DAPO-DeepSeek} checkpoints on all 14,042 MMLU test questions~\cite{hendrycks2020measuring}, using both the standard subject-wise 5-shot protocol and zero-shot constrained next-token scoring. We also evaluate on 7,080 clean HotpotQA validation questions after excluding the 325 SelectBench questions, in both fixed-context and closed-book modes. All dataset files, model exports, prompts, and trajectories are SHA-256 inventoried.

\subsection{Baselines}

We compare the original checkpoint, \textsc{DAPO-Rule}, and \textsc{DAPO-DeepSeek} using identical test prompts and a deterministic evaluator. Formal generation uses one trajectory per example with \texttt{do\_sample=false}, temperature $0$, top-$p$ $1$, no top-$k$ truncation, a $3{,}072$-token output limit, and seed $42$. The evaluator applies the Rule matching procedure described above to the visible answer for correctness, and to the complete final assistant turn for forbidden adoption or injection detection. This shared measurement regime enables paired comparisons across all three checkpoints without evaluating \textsc{DAPO-DeepSeek} using its training-time judge. Nevertheless, because the two trained checkpoints differ in prompt-batch size, GPU count, and execution history, their comparison is descriptive rather than a controlled reward-model ablation.

\subsection{Metrics}

Strict success requires a successful tool call, a correct and complete visible answer, and no detected forbidden adoption or injection following. We additionally report correct-and-complete answers, forbidden adoption, injection following, missing final answers, over-refusal, overlong output, tool success, and mean generated tokens. Injection following is conditioned on the 275 examples that contain an injection; all other metrics are computed over all 325 examples. From the paired per-example records, we compute percentile-bootstrap confidence intervals with 10,000 resamples and seed 42, and apply two-sided exact McNemar tests. We jointly apply Holm correction across nine prespecified comparisons: strict success, forbidden adoption, and injection following for each of the three model pairs. Each reward variant has one completed training and formal test run, so training-seed variability remains unmeasured.

\section{Results and Analysis}

\subsection{Main Results}

Table~\ref{tab:selectbench-main} reports the shared deterministic evaluation.
Strict success increases from 73/325 (22.46\%) for the original checkpoint to
83/325 (25.54\%) for \textsc{DAPO-Rule} and 86/325 (26.46\%) for
\textsc{DAPO-DeepSeek}.  The paired changes are $+3.08$ percentage points (20
improvements and 10 regressions) and $+4.00$ points (21 improvements and 8
regressions), respectively.  Their unadjusted 95\% bootstrap intervals are
$[0.00,6.46]$ and $[0.92,7.38]$ points.  The corresponding exact McNemar
$p$-values are 0.099 and 0.024, but after correction across the prespecified
code comparisons the Holm-adjusted values are 0.691 and 0.217.  We therefore interpret
the result as a modest directional improvement rather than a statistically
conclusive gain.  Tool success is 100\% for all three checkpoints.
\begin{table}[ht]
\centering
\small
\begin{tabular}{lrrrrrr}
\toprule
Model & Strict $\uparrow$ & Correct $\uparrow$ & Forbidden $\downarrow$ & $\Delta$ pp & 95\% CI & $p_{\mathrm{Holm}}$ \\
\midrule
Qwen3.5-4B & 73 (22.46\%) & 276 (84.92\%) & 236 (72.62\%) & --- & --- & --- \\
DAPO-Rule & 83 (25.54\%) & 276 (84.92\%) & 225 (69.23\%) & +3.08 & [0.00, 6.46] & 0.691 \\
DAPO-DeepSeek & 86 (26.46\%) & 282 (86.77\%) & 224 (68.92\%) & +4.00 & [0.92, 7.38] & 0.217 \\
\bottomrule
\end{tabular}
\caption{Primary deterministic results on the corrected 325-example SelectBench-v2 test set. Correct denotes a correct and complete visible answer. Confidence intervals are paired percentile-bootstrap intervals with 10,000 resamples; $p_{\mathrm{Holm}}$ is the exact McNemar $p$-value for strict success versus the original checkpoint after correction over the prespecified code-comparison family.}
\label{tab:selectbench-main}
\end{table}

Forbidden adoption falls from 236/325 (72.62\%) to 225/325 (69.23\%) with
\textsc{DAPO-Rule} and 224/325 (68.92\%) with
\textsc{DAPO-DeepSeek}.  The respective paired changes are $-3.38$ points
(95\% CI $[-7.08,0.31]$) and $-3.69$ points (95\% CI
$[-7.08,-0.31]$); neither remains significant after Holm correction.  Correct
and complete answers are unchanged at 84.92\% for \textsc{DAPO-Rule} and rise
to 86.77\% for \textsc{DAPO-DeepSeek}.  The large gap between factual
correctness and strict success is explained by their overlap: 203 original,
193 Rule, and 196 DeepSeek outputs are factually correct yet still adopt a
forbidden phrase.  Selective rejection of contamination, rather than factual
answering alone, is therefore the principal bottleneck under this evaluator.

Table~\ref{tab:selectbench-failures} shows that this shift does not extend to
prompt-injection resistance.  On the 275 eligible examples, injection
following changes from 14 (5.09\%) to 18 (6.55\%) and 16 (5.82\%); both
intervals include zero change.  By contrast, \textsc{DAPO-DeepSeek} reduces
over-refusal, missing final answers, and overlong responses.  Mean output
length also falls by about 12 generated tokens for both trained policies.
Overall, the completed evidence supports improved selective adoption on the
aggregate metric, but not uniform improvement across failure modes.

\begin{table}[ht]
\centering
\small
\begin{tabular}{lrrrrr}
\toprule
Model & Injection $\downarrow$ & Over-refusal $\downarrow$ & Missing final $\downarrow$ & Overlong $\downarrow$ & Mean tokens $\downarrow$ \\
\midrule
Qwen3.5-4B & 14 (5.09\%) & 3 (0.92\%) & 9 (2.77\%) & 7 (2.15\%) & 383.18 \\
DAPO-Rule & 18 (6.55\%) & 4 (1.23\%) & 9 (2.77\%) & 4 (1.23\%) & 371.20 \\
DAPO-DeepSeek & 16 (5.82\%) & 1 (0.31\%) & 5 (1.54\%) & 2 (0.62\%) & 371.32 \\
\bottomrule
\end{tabular}
\caption{Failure diagnostics on SelectBench-v2. Injection following is measured over the 275 injection-bearing examples; all other rates use 325 examples. Categories can overlap and therefore need not sum to the complement of strict success.}
\label{tab:selectbench-failures}
\end{table}

Table~\ref{tab:selectbench-pairwise} reports all nine comparisons in the
prespecified Holm family, including the Rule--DeepSeek contrast.  No comparison
remains significant after family-wise correction, and the direct difference
between the two reward variants is small on all three outcomes.

\begin{table}[ht]
\centering
\small
\begin{tabular}{llrrrr}
\toprule
Comparison & Outcome & $\Delta$ pp & 95\% CI & $p$ & $p_{\mathrm{Holm}}$ \\
\midrule
Original $\rightarrow$ Rule & Strict & +3.08 & [0.00, 6.46] & 0.099 & 0.691 \\
Original $\rightarrow$ DeepSeek & Strict & +4.00 & [0.92, 7.38] & 0.024 & 0.217 \\
Rule $\rightarrow$ DeepSeek & Strict & +0.92 & [$-1.85$, 3.69] & 0.664 & 1.000 \\
\midrule
Original $\rightarrow$ Rule & Forbidden & $-3.38$ & [$-7.08$, 0.31] & 0.099 & 0.691 \\
Original $\rightarrow$ DeepSeek & Forbidden & $-3.69$ & [$-7.08$, $-0.31$] & 0.050 & 0.401 \\
Rule $\rightarrow$ DeepSeek & Forbidden & $-0.31$ & [$-3.38$, 2.46] & 1.000 & 1.000 \\
\midrule
Original $\rightarrow$ Rule & Injection & +1.45 & [$-1.82$, 4.73] & 0.523 & 1.000 \\
Original $\rightarrow$ DeepSeek & Injection & +0.73 & [$-2.55$, 4.36] & 0.839 & 1.000 \\
Rule $\rightarrow$ DeepSeek & Injection & $-0.73$ & [$-3.64$, 2.18] & 0.804 & 1.000 \\
\bottomrule
\end{tabular}
\caption{All nine prespecified paired code comparisons on SelectBench-v2.  The
delta is the right model minus the left model in percentage points.  Intervals
are paired percentile-bootstrap 95\% CIs; $p$ is the two-sided exact McNemar
value and $p_{\mathrm{Holm}}$ corrects jointly over all nine rows.}
\label{tab:selectbench-pairwise}
\end{table}

\subsection{Evaluator and Dataset-Version Diagnostics}

The primary evaluator checks factual correctness only in the visible answer,
but scans the complete final assistant turn, including \texttt{<think>}, for
forbidden adoption.  It is consequently conservative when a forbidden phrase
is mentioned in reasoning, and it is aligned with the training signal for
\textsc{DAPO-Rule}.  Its advantage is that the same frozen decision rule and
the same 325 prompts are applied to all three checkpoints.  The semantic judge
used to train \textsc{DAPO-DeepSeek} is not used for the formal results.

Dataset versioning also materially affects measurement.  The original test
generator used a permissive supporting-fact fallback; the v2 audit replaces
only trusted evidence and changes the retrieval context in 233/325 examples,
while preserving questions, answers, and contamination.  All results in this
section are regenerated from v2.  Legacy v1 scores are excluded rather than
mixed with the corrected measurement regime.

\subsection{General Capability Retention}

On 14,042 MMLU questions, the original and \textsc{DAPO-DeepSeek} checkpoints
obtain 67.70 and 67.70 accuracy under 5-shot evaluation (a $+0.01$
percentage-point difference after rounding), with a paired interval spanning
zero and exact McNemar $p=1.000$.  Zero-shot accuracy changes from 58.65 to
58.85 ($+0.21$ points, 95\% CI $[0.06,0.36]$, $p=0.010$).  On the
7,080-question clean HotpotQA subset, fixed-context EM is identical at 60.03,
while F1 changes from 74.39 to 74.41.  Closed-book EM changes from 18.01 to
17.98 and F1 from 24.80 to 24.78.  The paired 95\% bootstrap intervals for both
HotpotQA modes include zero; Table~\ref{tab:generalization} reports the complete
paired intervals and the exact tests available for accuracy and EM.  These results show no material degradation on
the two measured general-capability evaluations for \textsc{DAPO-DeepSeek},
but do not establish capability preservation or broader robustness beyond them.

\begin{table}[ht]
\centering
\small
\begin{tabular}{lrrrrr}
\toprule
Evaluation & Original & DeepSeek & $\Delta$ pp & 95\% CI & $p$ \\
\midrule
MMLU 5-shot accuracy & 67.697 & 67.704 & +0.007 & [$-0.135$, 0.150] & 1.000 \\
MMLU zero-shot accuracy & 58.645 & 58.852 & +0.207 & [0.057, 0.356] & 0.010 \\
HotpotQA fixed-context EM & 60.028 & 60.028 & 0.000 & [$-0.184$, 0.184] & 1.000 \\
HotpotQA fixed-context F1 & 74.395 & 74.415 & +0.020 & [$-0.143$, 0.185] & --- \\
HotpotQA closed-book EM & 18.008 & 17.980 & $-0.028$ & [$-0.198$, 0.141] & 0.871 \\
HotpotQA closed-book F1 & 24.802 & 24.777 & $-0.024$ & [$-0.216$, 0.162] & --- \\
\bottomrule
\end{tabular}
\caption{General-capability retention for the original and DAPO-DeepSeek
checkpoints.  Values and differences are percentage points.  Confidence
intervals use 10,000 paired bootstrap resamples.  Reported $p$-values are
two-sided exact McNemar tests; they are not defined for token-overlap F1.
HotpotQA excludes all 325 SelectBench test questions.}
\label{tab:generalization}
\end{table}




\section{Related Work}

\subsection{Reinforcement Learning}

Reinforcement learning (RL) has become a cornerstone in aligning LLMs with human preferences and enhancing their reasoning capabilities. 
Proximal Policy Optimization (PPO) \cite{schulman2017ppo} is a widely used policy gradient algorithm that stabilizes training through clipped policy updates, but maintaining four models imposes heavy computational costs. Direct preference optimization (DPO) \cite{rafailov2023dpo} simplifies this by directly optimizing the preference gap between chosen and rejected responses, while Group Relative Policy Optimization (GRPO) \cite{shao2024deepseekmathgrpo} introduces group-based comparison to compute relative advantages across multiple outputs, enabling finer learning and strong results in LLM reasoning tasks such as mathematical problem solving.
With variants including DAPO \cite{yu2026dapo} and GSPO \cite{zheng2025gspo}, GRPO has become a new mainstream paradigm. 

Reinforcement learning extends beyond standard LLM post-training to enhance agent capabilities \cite{sha2025sem, chen2026research}. By designing appropriate reward functions and adapting optimization objectives to practical tasks, LLMs can learn to invoke external tools for search, file manipulation, and even complex GUI interactions \cite{zhang2026toolsneedunveilingtooluse}. 
However, reinforcement learning for agents faces distinct challenges, including sparse rewards arising from extended tool-use trajectories and heavy dependence on stable simulation environments. Nevertheless, these reinforcement learning methods remain essential for post-training optimization, ensuring models not only produce coherent text but also align with human values and exhibit stronger complex problem-solving abilities.





\subsection{Large Language Models as Agents}

Viewing large language models as autonomous agents capable of planning and executing multi-step reasoning has emerged as a new paradigm~\cite{shi2026surgent}. An agent can be decomposed into an LLM and a harness that comprises memory, planning, tools, and other components \cite{meng2026agent}. The LLM serves as the brain and core of the agent, iteratively refining tasks, retrieving information, and generating solutions under the orchestration of the harness \cite{yao2026harness}. However, the over-refusal problem inherent in LLMs persists when agents invoke retrieval tools. Our method explicitly trains the model through reinforcement learning to extract useful information from retrieved results that contain harmful content, thereby mitigating the underutilization of resources.

\section{Conclusion}

We introduced SelectBench and a DAPO post-training framework for selective
evidence adoption from retrieval contexts containing supported facts alongside
misleading or instruction-like text.  On the corrected SelectBench-v2 test set,
both reward variants improve strict success and reduce forbidden adoption, with
the DeepSeek-trained checkpoint also improving answer completeness and
retaining performance on MMLU and clean HotpotQA.  The gains are modest, do not
remain significant after multiple-comparison correction, and do not improve
prompt-injection following.  These results position selective evidence
adoption as a distinct and measurable post-training objective while showing
that stronger injection resistance, replicated training runs, and broader
retrieval settings remain necessary.

\bibliography{main}

@inproceedings{li2025hierarchical,
  title={Hierarchical safety realignment: Lightweight restoration of safety in pruned large vision-language models},
  author={Li, Yue and Yi, Xin and Shi, Dongsheng and De Melo, Gerard and Wang, Xiaoling and Wang, Linlin},
  booktitle={Findings of the Association for Computational Linguistics: ACL 2025},
  pages={7600--7612},
  year={2025}
}

@article{yi2025latent,
  title={Latent-space adversarial training with post-aware calibration for defending large language models against jailbreak attacks},
  author={Yi, Xin and Li, Yue and Shi, Dongsheng and Wang, Linlin and Wang, Xiaoling and He, Liang},
  journal={Expert Systems with Applications},
  pages={129101},
  year={2025},
  publisher={Elsevier}
}

@article{yi2025unifiededucation,
  title={Unified defense for large language models against jailbreak and fine-tuning attacks in education},
  author={Yi, Xin and Li, Yue and Shi, Dongsheng and Wang, Linlin and Wang, Xiaoling and He, Liang},
  journal={arXiv preprint arXiv:2511.14423},
  year={2025}
}

@article{li2025construction,
  title={From Construction to Injection: Edit-Based Fingerprints for Large Language Models},
  author={Li, Yue and Yi, Xin and Shi, Dongsheng and Cui, Yongyi and de Melo, Gerard and Wang, Linlin},
  journal={arXiv preprint arXiv:2509.03122},
  year={2025}
}

@article{yi2025unified,
  title={Unified attacks to large language model watermarks: spoofing and scrubbing in unauthorized knowledge distillation},
  author={Yi, Xin and Li, Yue and Zheng, Shunfan and Wang, Linlin and Wang, Xiaoling and He, Liang},
  journal={Knowledge-Based Systems},
  pages={114295},
  year={2025},
  publisher={Elsevier}
}

@misc{qwen35blog,
    title = {Qwen3.5: Accelerating Productivity with Native Multimodal Agents},
    url = {https://qwen.ai/blog?id=qwen3.5},
    author = {Qwen Team},
    month = {February},
    year = {2026}
}

@inproceedings{
chen2026research,
title={ReSearch: Learning to Reason with Search for {LLM}s via Reinforcement Learning},
author={Mingyang Chen and Linzhuang Sun and Tianpeng Li and sunhaoze and ZhouYijie and Chenzheng Zhu and Haofen Wang and Jeff Z. Pan and Wen Zhang and Huajun Chen and Fan Yang and Zenan Zhou and Weipeng Chen},
booktitle={The Thirty-ninth Annual Conference on Neural Information Processing Systems},
year={2026},
url={https://openreview.net/forum?id=OuGAwwAT8G}
}

@article{sha2025sem,
  title={Sem: Reinforcement learning for search-efficient large language models},
  author={Sha, Zeyang and Cui, Shiwen and Wang, Weiqiang},
  journal={arXiv preprint arXiv:2505.07903},
  year={2025}
}

@inproceedings{
zhang2025falsereject,
title={FalseReject: A Resource for Improving Contextual Safety and Mitigating Over-Refusals in {LLM}s via Structured Reasoning},
author={Zhehao Zhang and Weijie Xu and Fanyou Wu and Chandan K. Reddy},
booktitle={Second Conference on Language Modeling},
year={2025},
url={https://openreview.net/forum?id=1w9Hay7tvm}
}

@article{shi2026surgent,
  title={SURGENT: A Surgical Multi-Agent Assistance System Across the Perioperative Workflow},
  author={Shi, Dongsheng and Li, Yue and Yi, Xin and Cui, Yongyi and Feng, Huawei and Wang, Linlin},
  journal={arXiv preprint arXiv:2605.29368},
  year={2026}
}

@article{hendrycks2020measuring,
  title={Measuring massive multitask language understanding},
  author={Hendrycks, Dan and Burns, Collin and Basart, Steven and Zou, Andy and Mazeika, Mantas and Song, Dawn and Steinhardt, Jacob},
  journal={arXiv preprint arXiv:2009.03300},
  year={2020}
}

@misc{zhang2026toolsneedunveilingtooluse,
      title={Are Tools All We Need? Unveiling the Tool-Use Tax in LLM Agents}, 
      author={Kaituo Zhang and Zhen Xiong and Mingyu Zhong and Zhimeng Jiang and Zhouyuan Yuan and Zhecheng Li and Ying Lin},
      year={2026},
      eprint={2605.00136},
      archivePrefix={arXiv},
      primaryClass={cs.AI},
      url={https://arxiv.org/abs/2605.00136}, 
}

@article{
zhang2026a,
title={A Survey on Evaluating Quality and Trustworthiness in {LLM}-Generated Data},
author={Kaituo Zhang and Mingzhi Hu and Hoang Anh Duy Le and Fariha Kabir Torsha and Zhimeng Jiang and Minh Khai Bui and Chia-Yuan Chang and Yu-Neng Chuang and Zhen Xiong and Ying Lin and Guanchu Wang and Na Zou},
journal={Transactions on Machine Learning Research},
issn={2835-8856},
year={2026},
url={https://openreview.net/forum?id=f2gS9Ly6tA},
note={}
}

@article{shi2026benchmarking,
  title={Benchmarking large language models for end-to-end clinical support in traditional chinese medicine},
  author={Shi, Dongsheng and Yi, Xin and Li, Yue and Wang, Linlin},
  journal={Expert Systems with Applications},
  pages={132267},
  year={2026},
  publisher={Elsevier}
}

@article{li2026agmark,
  title={AGMark: Attention-Guided Dynamic Watermarking for Large Vision-Language Models},
  author={Li, Yue and Yi, Xin and Shi, Dongsheng and Cui, Yongyi and de Melo, Gerard and Wang, Linlin},
  journal={arXiv preprint arXiv:2602.09611},
  year={2026}
}

@article{yu2026dapo,
  title={Dapo: An open-source llm reinforcement learning system at scale},
  author={Yu, Qiying and Zhang, Zheng and Zhu, Ruofei and Yuan, Yufeng and Zuo, Xiaochen and Yue, Yu and Dai, Weinan and Fan, Tiantian and Liu, Gaohong and Liu, Lingjun and others},
  journal={Advances in Neural Information Processing Systems},
  volume={38},
  pages={113222--113244},
  year={2026}
}

@article{schulman2017ppo,
  title={Proximal policy optimization algorithms},
  author={Schulman, John and Wolski, Filip and Dhariwal, Prafulla and Radford, Alec and Klimov, Oleg},
  journal={arXiv preprint arXiv:1707.06347},
  year={2017}
}

@article{rafailov2023dpo,
  title={Direct preference optimization: Your language model is secretly a reward model},
  author={Rafailov, Rafael and Sharma, Archit and Mitchell, Eric and Manning, Christopher D and Ermon, Stefano and Finn, Chelsea},
  journal={Advances in neural information processing systems},
  volume={36},
  pages={53728--53741},
  year={2023}
}

@article{shao2024deepseekmathgrpo,
  title={Deepseekmath: Pushing the limits of mathematical reasoning in open language models},
  author={Shao, Zhihong and Wang, Peiyi and Zhu, Qihao and Xu, Runxin and Song, Junxiao and Bi, Xiao and Zhang, Haowei and Zhang, Mingchuan and Li, YK and Wu, Yang and others},
  journal={arXiv preprint arXiv:2402.03300},
  year={2024}
}

@article{zheng2025gspo,
  title={Group sequence policy optimization},
  author={Zheng, Chujie and Liu, Shixuan and Li, Mingze and Chen, Xiong-Hui and Yu, Bowen and Gao, Chang and Dang, Kai and Liu, Yuqiong and Men, Rui and Yang, An and others},
  journal={arXiv preprint arXiv:2507.18071},
  year={2025}
}

@article{yao2026harness,
  title={Harness-Bench: Measuring harness effects across models in realistic agent workflows},
  author={Yao, Yilun and Tan, Xinyu and Liu, Chao-Hsuan and Li, Yaoming and Wang, Zhengyang and Yu, Wenhan and Tan, Zhewen and Tian, Yuxuan and Zhao, Guangxiang and Sun, Lin and others},
  journal={arXiv preprint arXiv:2605.27922},
  year={2026}
}

@article{meng2026agent,
  title={Agent harness for large language model agents: A survey},
  author={Meng, Qianyu and Wang, Yanan and Chen, Liyi and Li, Yihang and Wu, Wei and Jiang, Wenyuan and Wang, Qimeng and Lu, Chengqiang and Gao, Yan and Wu, Yi and others},
  year={2026},
  publisher={Preprints}
}

@inproceedings{yang2018hotpotqa,
  title={HotpotQA: A Dataset for Diverse, Explainable Multi-hop Question Answering},
  author={Yang, Zhilin and Qi, Peng and Zhang, Saizheng and Bengio, Yoshua and Cohen, William W. and Salakhutdinov, Ruslan and Manning, Christopher D.},
  booktitle={Proceedings of the 2018 Conference on Empirical Methods in Natural Language Processing},
  pages={2369--2380},
  year={2018}
}

@inproceedings{ho2020constructing,
  title={Constructing A Multi-hop QA Dataset for Comprehensive Evaluation of Reasoning Steps},
  author={Ho, Xanh and Nguyen, Anh-Khoa Duong and Sugawara, Saku and Aizawa, Akiko},
  booktitle={Proceedings of the 28th International Conference on Computational Linguistics},
  pages={6609--6625},
  year={2020}
}

@article{trivedi2022musique,
  title={MuSiQue: Multihop Questions via Single-hop Question Composition},
  author={Trivedi, Harsh and Balasubramanian, Niranjan and Khot, Tushar and Sabharwal, Ashish},
  journal={Transactions of the Association for Computational Linguistics},
  volume={10},
  pages={539--554},
  year={2022}
}

@article{xu2026deepseek,
  title={Deepseek-v4: Towards highly efficient million-token context intelligence},
  author={Xu, Anyi and Lin, Bangcai and Xue, Bing and Wang, Bingxuan and Xu, Bingzheng and Wu, Bochao and Zhang, Bowei and Lin, Chaofan and Dong, Chen and Ling, Chenchen and others},
  journal={arXiv preprint arXiv:2606.19348},
  year={2026}
}
\bibliographystyle{iclr2026_conference}

\appendix







\end{document}